\documentclass{styles/svproc}
\usepackage{url}

\usepackage{graphicx}
\usepackage{multicol}
\usepackage[bottom]{footmisc}
\usepackage{todonotes}
\usepackage{amsfonts}
\usepackage{amsmath}
\usepackage{times}
\usepackage{graphicx}
\usepackage[caption=false]{subfig}
\usepackage[ruled, english]{algorithm2e}
\usepackage{tabularx}
\usepackage{verbatim}
\usepackage{mathtools}
\mathtoolsset{showonlyrefs}
\usepackage{hyperref}
\usepackage{xcolor}

\usepackage{booktabs}

\begin{document}
\mainmatter              
\title{Event-based Visual Tracking in Dynamic Environments \footnote{\color{red}This preprint has not undergone peer review or any post-submission improvements or
corrections. The Version of Record of this contribution is published in ROBOT2022: Fifth Iberian Robotics Conference and is available online at {\tt\small \url{https://doi.org/10.1007/978-3-031-21065-5\_15}}.}}
\titlerunning{Event-based Visual Tracking in Dynamic Environments}  
\toctitle{Event-based Visual Tracking in Dynamic Environments}
\author{{Irene Perez-Salesa \and Rodrigo Aldana-López \and Carlos Sagüés}}
\authorrunning{{Irene Perez-Salesa et al.}} 
%
\tocauthor{{Irene Perez-Salesa, Rodrigo Aldana-López, Carlos Sagüés}}
\institute{{Universidad de Zaragoza, Maria de Luna 1, 50018 Zaragoza, Spain,}\\
\email{{i.perez@unizar.es, rodrigo.aldana.lopez@gmail.com, csagues@unizar.es}}}%

\maketitle              

\begin{abstract}
Visual object tracking under challenging conditions of motion and light can be hindered by the capabilities of conventional cameras, prone to producing images with motion blur. Event cameras are novel sensors suited to robustly perform vision tasks under these conditions. However, due to the nature of their output, applying them to object detection and tracking is non-trivial. In this work, we propose a framework to take advantage of both event cameras and off-the-shelf deep learning for object tracking. We show that reconstructing event data into intensity frames improves the tracking performance in conditions under which conventional cameras fail to provide acceptable results.
\end{abstract}

\section{Introduction}

Object tracking is a widely studied problem in the field of robotics. With the development of neural networks which can perform object detection in real-time, trackers-by-detection have become of interest. The use of neural networks for detection provides new application opportunities, due to their flexibility in recognition of objects and the lack of markers attached to the targets. However, the development of robotic applications featuring highly dynamic environments and other challenging aspects needs vision sensors able to produce clear images in these conditions. Conventional cameras are prone to producing images affected by motion blur or exposure issues, which can cause the failure of the detection task. Other requirements such as a high frequency of the information might be needed as well. 

Event cameras are novel vision sensors that show great potential for robotic applications. They asynchronously capture light intensity changes in each independent pixel, rather than a full image of absolute intensity, resulting in a stream of events as the output. Due to their low latency, high dynamic range and robustness to motion blur, they are able to capture more information in challenging conditions of motion and light where conventional cameras fail to produce clear images. Additionally, these sensors reduce communication bandwidth and storage compared to conventional cameras since information is only transmitted when events occur \cite{gallego2022}. Despite this, the use of event cameras for computer vision applications is not trivial, given the nature of their output. Note that events are caused by the relative motion between the elements of the scene and the camera. Thus, extracting relevant information is challenging, especially when the camera moves, since events are caused by both the target and the background. Moreover, traditional vision algorithms generally take absolute intensity images as an input, so an adaptation is needed to use them on events.

Several works have tried to apply deep learning for object detection and tracking tasks on event data. One approach relies on conventional frames taken simultaneously with the stream of events. In \cite{Dubeau}, \cite{Zhang2021}, \cite{Wang2021}, events are fed alongside frames to a convolutional neural network (CNN) to identify the object using information from both domains. Event data is used in \cite{Liu2016} to speed up the performance of the frame-based tracker, applying a CNN only on regions of interest located using events. In \cite{Jiang2021}, inter-frame events are assigned to bounding boxes detected on the frames and used to compute optical flow. This combined approach improves the performance of frame-based trackers, but relying solely on conventional frames for the detection task might fail when the images show issues like motion blur, and events often need to be grouped according to the frequency of the frames. 

Other works have adapted frame-based detection neural networks to take event data as the only input. A tailor-made model is proposed in \cite{Perot}, while in \cite{Iacono2018}, \cite{Jiang2020} existing architectures are retrained. In these works, events are grouped into temporal windows and arranged into pseudo-images that show a dense representation of events. However, the representation of an object in the event domain can change greatly according to the temporal window and the amount of events happening at every point in time. In \cite{Perot}, \cite{Iacono2018}, the windows used are in the range of 30 to 100 ms, meaning that information would be obtained at a slower frequency than with conventional cameras. In addition, the temporal window has to be taken into account during training, as it is pointed out in \cite{Iacono2018}.

Finally, an issue for both of these approaches is the scarcity of readily available, large labelled event datasets to train the networks for object detection. This also makes comparisons harder, since each approach is tested on an ad hoc dataset.

Recent works have proposed to reconstruct absolute intensity images from event data with neural networks. In \cite{Rebecq2019}, the E2VID neural network is able to produce a sequence of grayscale frames from an event stream, by using recurrent convolutional layers to remember relevant past information from the events. The reconstructed frames are tested for object classification and visual-inertial odometry, showing its usefulness for these tasks. In \cite{Scheerlinck2020}, FireNet is proposed based on E2VID, with a smaller size allowing it to run faster while producing frames of comparable quality in most cases. Despite this, to the best of our knowledge, this method has not been studied in the context of object tracking. 

Motivated by this discussion, we propose an event-based tracker-by-detection, which reconstructs intensity frames from events and applies a frame-based neural network detector, followed by a multi-rate state estimation stage. As contribution, we show that by dividing the problem of \emph{detection on events} into \emph{image reconstruction from events} and \emph{detection on intensity frames}, we can exploit the advantage in perception of the event camera while bypassing common issues of event-based approaches, such as the need for labelled event data and the effect of the event window at the training stage of the detector. The tracker is tested on scenes with ego-motion of the camera and challenging object motion and light conditions, where conventional cameras typically perform poorly. The tracker produces estimates at high frequency at an asynchronous rate, adapted to the dynamic characteristics of the scene. 

\begin{figure}[t]
         \centering
         \includegraphics[width=\textwidth]{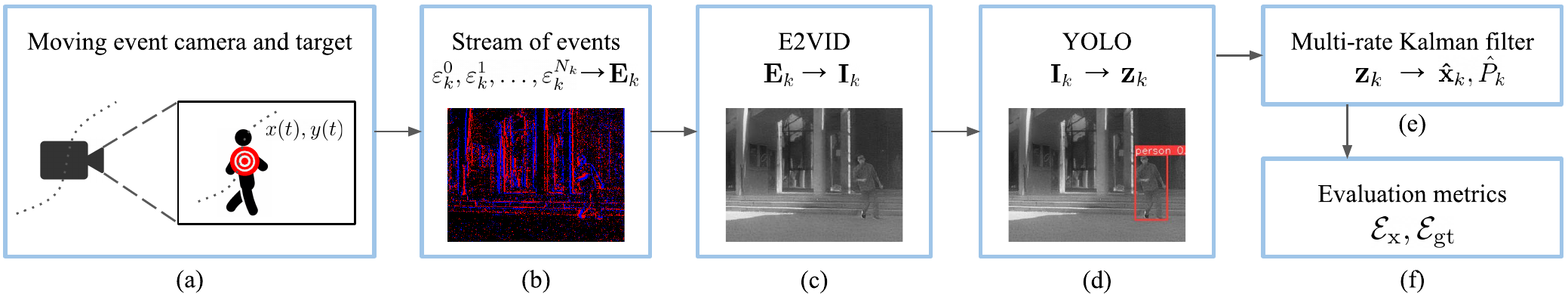}
        \caption{\textbf{(a)} Proposed framework for target tracking using a moving event camera. \textbf{(b)} The sensor captures a stream of events, that is partitioned into tensors. \textbf{(c)} An E2VID network reconstructs intensity frames from the event tensors. \textbf{(d)} A YOLO detector produces a measurement of the target position that is fed to a multi-rate Kalman filter in order to estimate the target's state \textbf{(e)}. Finally, the results are evaluated \textbf{(f)}.} 
        \label{fig:pipeline}
\end{figure}

\section{Framework}\label{framework}
 
\subsection{Problem Statement and Summary of the Proposal}
In this work, we study the following problem. Consider a scene containing targets of interest, an event camera, and possibly additional dynamic elements in the scene not related to the target, as depicted in Fig. \ref{fig:pipeline}-(a). Moreover, we are interested in situations in which both the targets and the camera move, so events are caused by the background elements as well as the relevant targets. For simplicity, we now define the parameters of the problem and our framework for a single target. Nevertheless, our approach can be extended to multiple object tracking in the same way as traditional frame-based trackers-by-detection, by implementing association and re-identification strategies accordingly. Let the position of the target projected into the image plane of the event camera be $[x(t),y(t)]^T\in\mathbb{R}^2$ for $t$ in an interval of interest $[0,T_f]$. Motivated by mobile robot tracking tasks, we want to obtain an estimate of the target position $[x(t), y(t)]^T$ as well as its velocity $[v_x(t), v_y(t)]^T$ for any $t\in[0,T_f]$ by an online object tracking strategy. This is, the estimates obtained at $t\in[0,T_f]$ are produced using events recorded before $t$. 
As described in Fig. \ref{fig:pipeline}, our proposal uses an image reconstruction neural network fed by events to create absolute intensity frames. In a later step, an object detection network is used to locate the object within the reconstructed frame. Such detection is used as a measurement in a multi-rate Kalman filter in order to obtain a full estimate of the state of the target with a measure of its uncertainty.

\subsection{Image Reconstruction}

Similar to \cite{Rebecq2019}, the asynchronous stream of events is processed by creating consecutive spatio-temporal voxel grids. The stream of events is partitioned into sequential groups of events $\mathbf{e}_k = \{\varepsilon_k^i\}_{i=0}^{N_k}$ where $\varepsilon^i_k$ contains a timestamp $t_k^i$, the pixel coordinate and polarity of the recorded event. Note that a group of events $\mathbf{e}_k$ spans a time interval of length $\mathrm{\Delta} T_k= t_{k}^{N_k-1} - t_{k}^0 $ that may not be constant. As described in more detail in \cite{Rebecq2019}, the group of events $\mathbf{e}_k$ is converted into an event tensor $\mathbf{E}_k$.
We group events into windows of a fixed amount of events, i.e. a fixed value of $N_k=N, \forall k\geq 0$ is chosen. 
Hence, given that events represent changes and motion in the scene, we create a new reconstruction only when a sufficient amount of events has happened. Therefore, no redundant images are produced regardless of the time that has passed. We find that this method provides the best results, in contrast to producing reconstructed frames at a fixed temporal rate, since it allows the frequency of the reconstruction to adapt to the dynamics of the scene.
In order to make the event window size agnostic to sensor dimensions, we refer to it by its number of events per pixel $n$, which is the number of events per window $N$ divided by the sensor size.

The information registered in the event tensor $\mathbf{E}_k$ is reconstructed into an absolute intensity image $\mathbf{I}_k$ by using E2VID, a recurrent neural network that takes event data as an input and produces grayscale frames. We propose this network over FireNet \cite{Scheerlinck2020} for its better quality when reconstructing scenes under fast motions, in order to preserve a good performance in highly dynamic environments. The time instant associated with $\mathbf{E}_k$ is the timestamp of its last registered event, denoted as $t_k:=t_{k}^{N_k -1}$. Therefore, the time interval between two consecutive reconstructions, $\mathbf{I}_{k}$ and $\mathbf{I}_{k+1}$, is given by $\mathrm{\Delta} t_k = t_{k+1} - t_{k}$.

\subsection{Object Detection}

A frame-based detection neural network has been used on the reconstructed frames, in order to obtain a detection measurement $\mathbf{z}_k$ of the coordinates of the target in the reconstructed frame $\mathbf{I}_k$. The YOLOv5 network \cite{yolov5} has been chosen for its fast and real-time detection. By using this method, the detection of the target can be achieved without being restricted to specific settings and easily recognizable targets. We use an available pre-trained model of the YOLO network, with no adaptation for this particular case. Note that the detector is trained on conventinal frames, which makes its training independent of the event window used afterwards. That being said, event cameras typically have low resolution, so the performance of the detection network could be further improved by training it on low resolution frames.

The measurement $\mathbf{z}_k$ is computed as the image coordinates at the center of each detected object bounding box. Ideally, the detector would produce a correct detection of the object on every frame in the sequence. However, false positives and missed detections can be produced. In case of having more than one candidate bounding box for the same frame, the one with the highest confidence score is selected. 

\subsection{Tracker} 

The goal is to estimate $\mathbf{x}(t)=[x(t),y(t),v_x(t),v_y(t)]^T$ for all $t\in[0,T_f]$, i.e. during the whole experiment and not only when measurements are available. Thus, we model the double integrator continuous-time dynamics  as a sampled-data system \cite[Chapter 4.5]{Soderstrom2002}:
\begin{equation}\label{eq:dynamic-continous}
\begin{aligned}
	&\mathbf{x}(t^+) = F(t^+-t)\mathbf{x}(t) + G(t^+-t)\mathbf{w}(t), \;\; t^+\geq t \\
	&F(\tau) :=
    \begin{bmatrix}
    1 & 0 & \tau & 0 \\ 
    0 & 1 & 0 & \tau \\
    0 & 0 & 1 & 0 \\
    0 & 0 & 0 & 1 \\
    \end{bmatrix}, \;\;
    G(\tau) :=
    \begin{bmatrix}
    \tau^2/2 & 0 \\
    0 & \tau^2/2 \\
    \tau & 0 \\
    0 & \tau \\
    \end{bmatrix}
\end{aligned}
\end{equation}
where $\mathbf{w}(t)=[a_x(t),a_y(t)]^T$ is a normal random variable with $\text{cov}\{\mathbf{w}(t)\}=Q\in\mathbb{R}^{2\times 2}$ modeling unknown inputs, disturbances and non-modeled dynamics. In addition, assume that $\mathbf{x}(0)$ is normally distributed with mean $\hat{\mathbf{x}}_0$ and covariance $\hat{P}_0$. Hence, letting $\mathbf{x}_k:=\mathbf{x}(t_k), \mathbf{w}_k:=\mathbf{w}(t_k)$, $F_k:=F(\mathrm{\Delta} t_k), G_k:=G(\mathrm{\Delta} t_k)$, as well as $t=t_k$ and $t^+=t_{k+1}$, a discrete-time model is obtained from \eqref{eq:dynamic-continous}:
\begin{equation}\label{eq:dynamic-discrete}
\begin{aligned}
	&\mathbf{x}_{k+1} = F_{k}\mathbf{x}_{k} + G_{k}\mathbf{w}_k, \;\; k \geq 0 
\end{aligned}
\end{equation}
Additionally, we consider the detection measurements $\mathbf{z}_k$ to be modelled as
\begin{equation}
\begin{aligned}
\mathbf{z}_k = H\mathbf{x}_k + \mathbf{v}_k, \;\;
H =   \begin{bmatrix}
		1 & 0 & 0 & 0\\	
		0 & 1 & 0 & 0\\		
		\end{bmatrix}
\end{aligned}
\end{equation}
where $\mathbf{v}_k$ is the measurement noise following a normal distribution with zero mean and $\text{cov}\{\mathbf{v}_k\}=R_k$ as an abstraction of the quality for the detection framework.  Using this model, a multi-rate Kalman filter is used to construct a causal estimate $\hat{\mathbf{x}}_k=\mathbb{E}\{\mathbf{x}_k|\mathbf{z}_0,\dots,\mathbf{z}_k\}$ for the process $\{\mathbf{x}_k\}_{k\geq 0}$ with error covariance $\hat{P}_k = \mathbb{E}\{({\mathbf{x}}_k-\hat{\mathbf{x}}_k)({\mathbf{x}}_k-\hat{\mathbf{x}}_k)^T\}$. As usual, the estimate $\mathbf{\hat{x}}_k$ is computed recursively by using prediction and measurement update stages \cite{Kalman1960}:
\begin{equation}
\label{eq:kalman}
\begin{aligned}
\begin{array}{ll}
    \mathbf{\hat{x}}_{k | k-1} &= F_k\mathbf{\hat{x}}_{k-1} \\
    \hat{P}_{k|k-1} &= F_k \hat{P}_{k-1} F_k^T + G_kQG_k^T \\
    L_k &= \hat{P}_{k | k -1}H^T (H\hat{P}_{k|k-1}H^T + R_k)^{-1}\\
    \mathbf{\hat{x}}_k &= \mathbf{\hat{x}}_{k | k-1} + L_k(\mathbf{z}_k - H\mathbf{\hat{x}}_{k | k-1}) \\
    \hat{P}_k &= (I - L_k H)\hat{P}_{k|k-1}\\
\end{array}
\end{aligned}
\end{equation}

Since a new measurement is obtained only when a certain amount of events $N_k=N$ is reached, the tracker updates its estimate asynchronously. Similarly, an estimate $\hat{\mathbf{x}}(t)=\mathbb{E}\{\mathbf{x}(t)|\mathbf{z}_0,\dots,\mathbf{z}_k\}$ with $\hat{P}(t) = \mathbb{E}\{({\mathbf{x}}(t)-\hat{\mathbf{x}}(t))({\mathbf{x}}(t)-\hat{\mathbf{x}}(t))^T\}$ is available for all $t\in[t_k,t_{k+1})$ as
\begin{equation}
\label{eq:kalman_t}
\begin{aligned}
    \mathbf{\hat{x}}(t)&= F(t-t_k)\mathbf{\hat{x}}_{k} \\
    \hat{P}(t) &= F(t-t_k)\hat{P}_kF(t-t_k)^T + G(t-t_k)QG(t-t_k)^T
\end{aligned}
\end{equation}

\begin{algorithm}[b]
        \caption{Event-based tracking algorithm}\label{alg_estim}
        $k \gets 0$\\
        \SetKwRepeat{Repeat}{For any}{end}
        \Repeat($\ t\in[0,T_f]$){}{
            {
            \eIf{$N$ new events were captured}{
            Construct tensor $\mathbf{E}_k$ from events $\mathbf{e}_k$; Reconstruct $\mathbf{I}_k$ from $\mathbf{E}_k$; Detect $\mathbf{z}_k$ on $\mathbf{I}_k$; 
            Compute $\mathbf{\hat{x}}_k$ using $\mathbf{\hat{x}}_{k-1}$ and $\mathbf{z}_k$ as in \eqref{eq:kalman}; $k \gets k + 1$, $t_k\gets t$
            }
            {
                Compute $\mathbf{\hat{x}}(t)$ using $\mathbf{\hat{x}}_k$ for $t\geq t_k$ as in  \eqref{eq:kalman_t}
            }
            }
        }
    \end{algorithm}

Algorithm \ref{alg_estim} summarizes the steps required to implement the proposed pipeline. 

    
\subsection{Evaluation Metrics}

In order to evaluate our event-based tracker versus a baseline conventional frame-based one, we establish some metrics to measure performance when estimating the state of the target object within $t \in [0,T_f]$. First, note that an analogous estimation for $\mathbf{x}(t)$ can be performed for the conventional frame-based setup by using \eqref{eq:kalman}, \eqref{eq:kalman_t} and setting $\mathrm{\Delta} t_k = 1/f_r$, where $f_r$ is the fixed frame rate of the camera. The goal of these metrics is to capture the fact that estimations for $t\in[0,T_f]$ must be obtained, whose uncertainty may increase between measurements. In addition, we evaluate the performance for any $t\in[0,T_f]$ for the sake of fairness, since estimations for our tracker and the baseline may not fall in the same time instants.
Hence, we first evaluate the confidence of the estimator under the assumption that the target follows \eqref{eq:dynamic-continous}, such that the estimate \eqref{eq:kalman_t} is unbiased. In this case, we evaluate the expected accumulated mean-squared error in an interval $[t,t^+], 0\leq t\leq t^+\leq T_f$ as:
\begin{equation}\label{eq:error-cov}
\begin{aligned}
\mathcal{E}_{\mathbf{x}}(t,t^+) &= \sqrt{\mathbb{E}\left\{\frac{1}{t^+-t}\int_t^{t^+}\|\mathbf{x}(\tau)-\mathbf{\hat{x}}(\tau)\|^2\text{d}\tau\right\}} =\sqrt{\frac{1}{t^+-t}\int_t^{t^+} \text{tr}(\hat{P}(\tau))\text{d}\tau}
\end{aligned}
\end{equation}
In practice, the target might not follow the model in \eqref{eq:dynamic-continous} exactly, since Gaussian assumptions on $\mathbf{v}_k, \mathbf{w}(t)$ may not hold. However, in some cases an unbiased groundtruth signal $\mathbf{z}_{\text{gt}}(t) = \mathbb{E}\{H\mathbf{x}(t)\}$ is available for evaluation. In these cases, we can evaluate the estimation with the following as well:
\begin{equation}\label{eq:error-x}
		\mathcal{E}_{\text{gt}}(t,t^+) = \sqrt{\frac{1}{t^+-t}\int_t^{t^+}\|\mathbf{z}_{\text{gt}}(\tau) - H\hat{\mathbf{x}}(\tau)\|^2\text{d}\tau}
\end{equation}
Note that $\mathcal{E}_{\mathbf{x}}(0,T_f), \mathcal{E}_{\text{gt}}(0,T_f)$ evaluate the performance during the whole interval of interest $[0,T_f]$. However, when comparing a conventional frame-based estimation and an event-based one, the first measurement might not be available for both of them at the same time, due to the different frame-rates used and to possible missed detections. For the sake of fairness, it is beneficial to evaluate $\mathcal{E}_{\mathbf{x}}(T_s,T_f), \mathcal{E}_{\text{gt}}(T_s,T_f)$ after some time $0\leq T_s\leq T_f$.

\section{Experiments}
The experiments have been performed on sequences from the VisEvent and Event Camera datasets \cite{Wang2021}, \cite{Mueggler}, which provide paired conventional frames and event data captured with a DAVIS camera, as well as groundtruth annotations for the frames in the VisEvent dataset. The sequences include challenging conditions such as fast motions, low light and high dynamic range. They also feature background events due to camera motion. 
First, we test the image reconstruction. We then evaluate the object detection task on the frames reconstructed from events. Lastly, we show tracking experiments by using the full pipeline proposed in Sect. \ref{framework}. The following experiments can be reproduced by using the code available at {\tt\small \url{https://github.com/ireneperezsalesa/event_tracking}}. 

\subsection{Image Reconstruction}

\begin{figure}[t]
  \subfloat{\includegraphics[width=0.19\columnwidth]{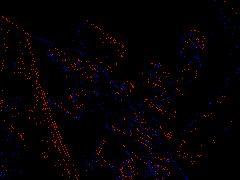}}\hfill%
  \subfloat{\includegraphics[width=0.19\columnwidth]{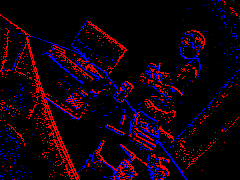}}\hfill%
  \subfloat{\includegraphics[width=0.19\columnwidth]{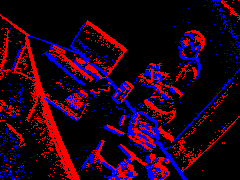}}\hfill%
  \subfloat{\includegraphics[width=0.19\columnwidth]{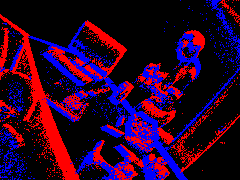}}\hfill%
  \subfloat{\includegraphics[width=0.19\columnwidth]{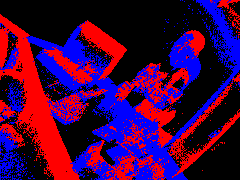}}\hfill%

  \subfloat[1000]{\includegraphics[width=0.19\columnwidth]{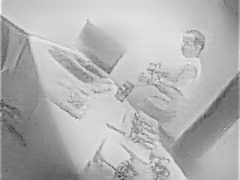}}\hfill%
  \subfloat[5000]{\includegraphics[width=0.19\columnwidth]{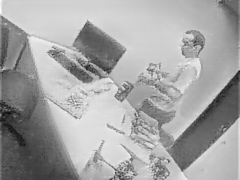}}\hfill%
  \subfloat[10000]{\includegraphics[width=0.19\columnwidth]{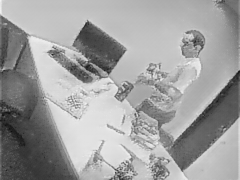}}\hfill%
  \subfloat[20000]{\includegraphics[width=0.19\columnwidth]{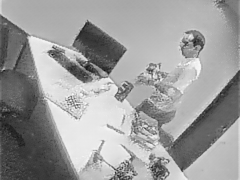}}\hfill%
  \subfloat[50000]{\includegraphics[width=0.19\columnwidth]{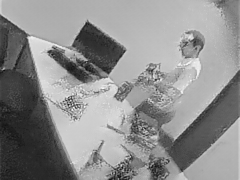}}\hfill%
     
    \caption{Event representation (\textit{top}) and reconstructed frames (\textit{bottom}) with different values of $N$. The reconstructed frame remains clear for a wide range of values.} 
    \label{fig:event-window}
\end{figure}

Recall that reconstructed frames are created from event data by grouping events into windows of a fixed event size $N$. This parameter influences the quality of the reconstructed frames, thus affecting the performance of the algorithms that use these frames as their input. Figure \ref{fig:event-window} shows a comparison of the event representation and the reconstructed frames obtained with different values of $N$. When the chosen value is extremely small, the image becomes faded, while some blurriness or warping may occur for values that are too large. However, it can be seen that, for a wide range of values of event window sizes, the reconstructed frame remains similar in quality. Therefore, $N$ does not need a very precise adjustment to ensure a reconstruction that properly captures the scene. Additionally, the representation of objects in the domain of absolute intensity is less affected by the window size than in the event domain. In particular, the reconstructed frame may retain more information for small windows. 

\subsection{Detection}\label{exp:detection}

\begin{figure}[t]
  \subfloat[]{\includegraphics[width=0.24\columnwidth]{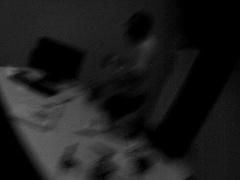}}\hfill%
  \subfloat[]{\includegraphics[width=0.24\columnwidth]{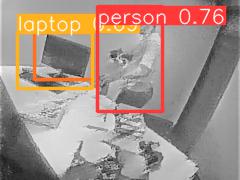}}\hfill%
  \subfloat[]{\includegraphics[width=0.24\columnwidth]{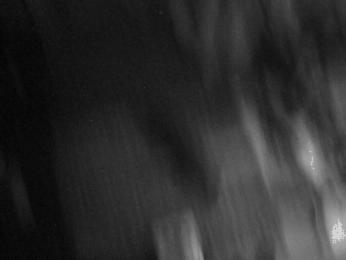}}\hfill
  \subfloat[]{\includegraphics[width=0.24\columnwidth]{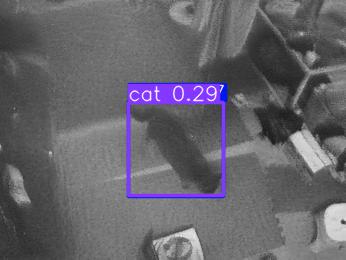}}\hfill

    \caption{Comparison of detection with conventional frames \textbf{(a, c)} and reconstructed ones \textbf{(b, d)}. The latter show clearer images, allowing detection of objects that are unrecognizable on the first.}
    \label{fig:conv-vs-event}
\end{figure}

\begin{figure}[t]
\centering
  \subfloat[No motion blur, class=`person'\label{fig:quality-person}]{\includegraphics[width=0.45\textwidth]{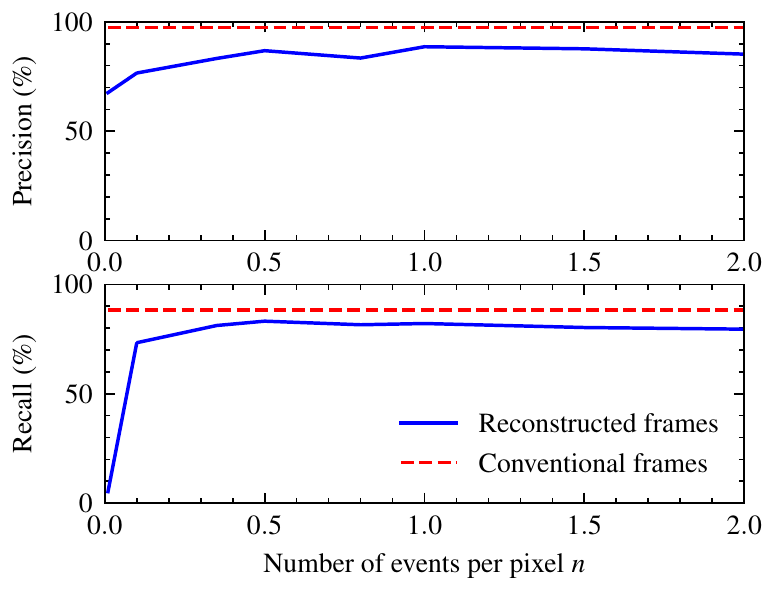}}\hfill%
  \subfloat[Motion blur, class=`cat'\label{fig:quality-cat}]{\includegraphics[width=0.45\textwidth]{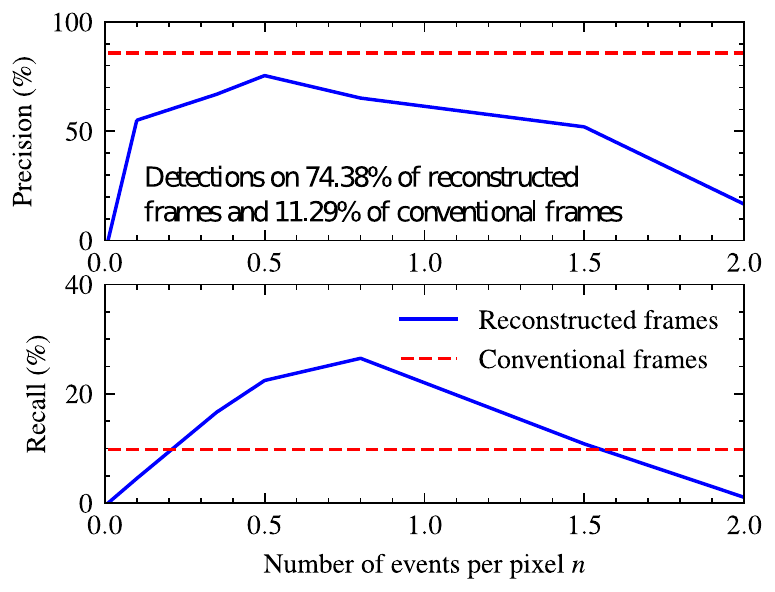}}%

    \caption{Detection quality at different values of event windows for the reconstructed frames, compared to baseline detection on conventional ones. Compared to conventional frames with no motion blur, the reconstructed frames show a slightly lower detection quality. When motion blur is present in conventional frames, the reconstructed ones are able to show a clearer image that allows to detect the object, achieving a much higher recall rate. Precision is still lower for reconstructed frames, but note that this metric is computed only on the images with found detections.} 
    \label{fig:quality}
\end{figure}

\begin{figure}[t]
         \centering
         \includegraphics[width=0.45\textwidth]{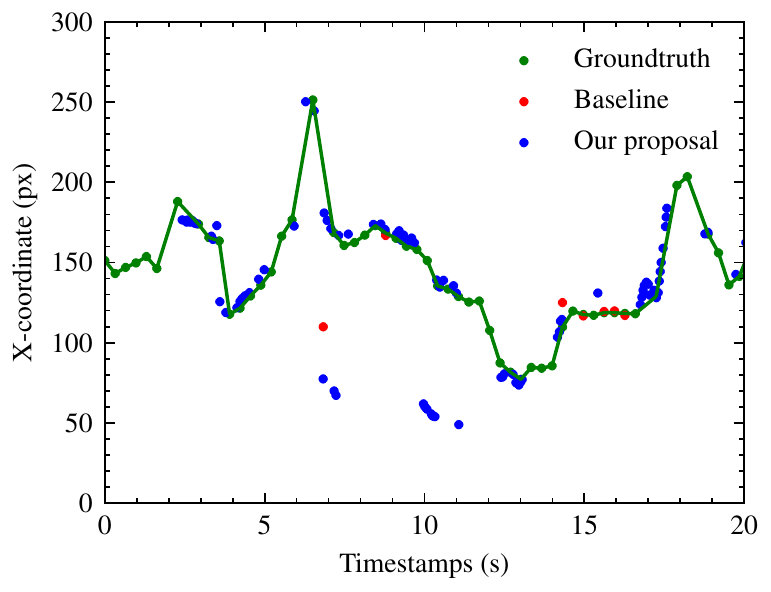}
     
        \caption{Detections with the baseline method and our proposal for a temporal sequence with motion blur. Our proposal provides more information about the location of the target: there are 83 correct measurements, while the baseline produces 6.} 
        \label{fig:dets}
\end{figure}

The performance of the object detection network on the reconstructed frames is also evaluated. Event cameras are expected to perform better than conventional ones on scenes with fast motions and challenging light conditions as in the examples shown in Fig. \ref{fig:conv-vs-event}. Note that reconstructed frames allow the detector to identify the objects that have become indistinguishable on the conventional ones. 
To compare the quality of detection with YOLO on reconstructed frames to a baseline detection on conventional ones, precision and recall metrics were computed as 
\begin{equation}
\label{eq:detection-metrics}
\begin{aligned}
\textsf{Precision} = \frac{\textsf{TP}}{\textsf{TP}+\textsf{FP}}, \;\; 
\textsf{Recall} = \frac{\textsf{TP}}{\textsf{TP}+\textsf{FN}} 
\end{aligned}
\end{equation}
where $\textsf{TP}$ is the number of true positives, $\textsf{FP}$ the amount of false positives and $\textsf{FN}$ the number of false negative detections. A true positive is considered for detections of the correct class and Intersection over Union $\textsf{IoU}>$ 50\%. Since the timestamps of the reconstructed frames don't necessarily match the groundtruth annotations, the bounding box labels have been interpolated. In cases where the dataset provides RGB conventional frames, they have been converted to grayscale for a fair comparison. As events can be grouped into windows of different sizes to produce reconstructions, the metrics were computed for several values of events per pixel $n$, in order to check its influence on the detection task. 
It is worth noting that the error in detection is partly due to the YOLO network itself. The quality of the detection varies with the object class, especially given that we are working with low resolution, grayscale images. For this reason, we are comparing the detection on the reconstructed frames to a baseline detection with the YOLO network on grayscale conventional frames of the same resolution.

Figure \ref{fig:quality-person} shows the precision and recall values on sequences that show little to no motion blur. Generally, reconstructed frames are not as sharp as conventional frames. Thus, the quality of detection on the reconstructed frames is slightly lower than that of conventional ones for this case. The detection quality metrics have also been computed on a sequence that shows strong motion blur on the conventional frames. Figure \ref{fig:quality-cat} shows that the reconstructed frames are then able to achieve much higher recall values than conventional ones, i.e. less missed detections are produced due to the improvement in image quality. The precision stays lower for the reconstructed frames, since this metric only takes into account the images for which detections are produced. In the case of conventional frames, detections are found on 11.29\% of the images, the ones with negligible motion blur. In contrast, for reconstructed images, detections are achieved on 74.38\% of the frames, even though they may contain some false positives. Note that the quality of detection drops for event window sizes for which suboptimal reconstructed frames are produced. 
Additionally, the asynchronous nature of events allows to produce reconstructed frames at higher frequencies than conventional cameras, meaning that a larger amount of information about the position of the target can be obtained for the same temporal sequence. Figure \ref{fig:dets} shows the detections on a sequence with strong motion blur on most of the conventional frames. Even though some false positives are obtained with our proposal, the amount of true positives is vastly superior than for the baseline detection, which provides no information for a large amount of time.

\subsection{Tracking}

The proposed tracker has been tested to estimate the position of a target on several sequences with camera motion, and its performance has been compared to a baseline tracker that uses the conventional frames provided by the dataset. The annotations included in the VisEvent dataset have been used to produce the groundtruth $\mathbf{z}_\text{gt}(t)$, by interpolating the center position of the bounding boxes to obtain values for the whole interval of interest $t \in [0, T_f]$. 

\begin{figure}[t]
\centering
  \subfloat[X-Coordinate]{\includegraphics[width=0.43\textwidth]{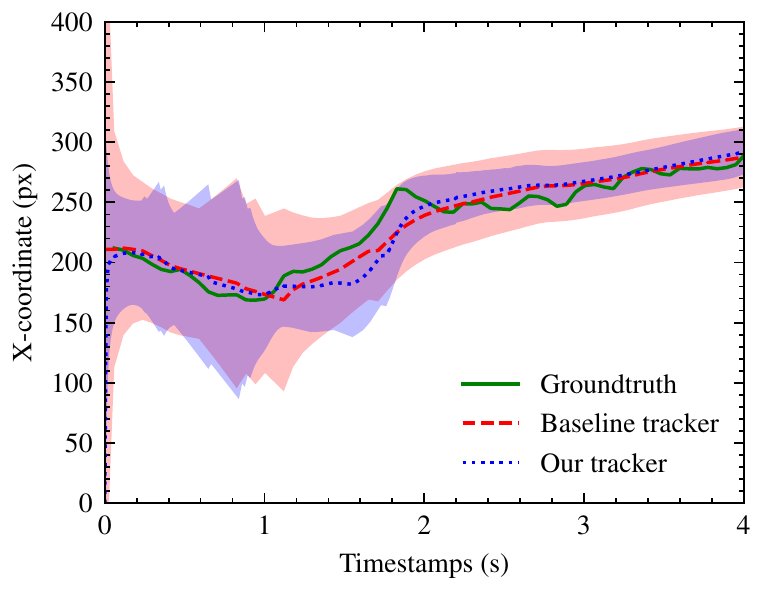}}\hfill%
  \subfloat[Y-Coordinate]{\includegraphics[width=0.43\textwidth]{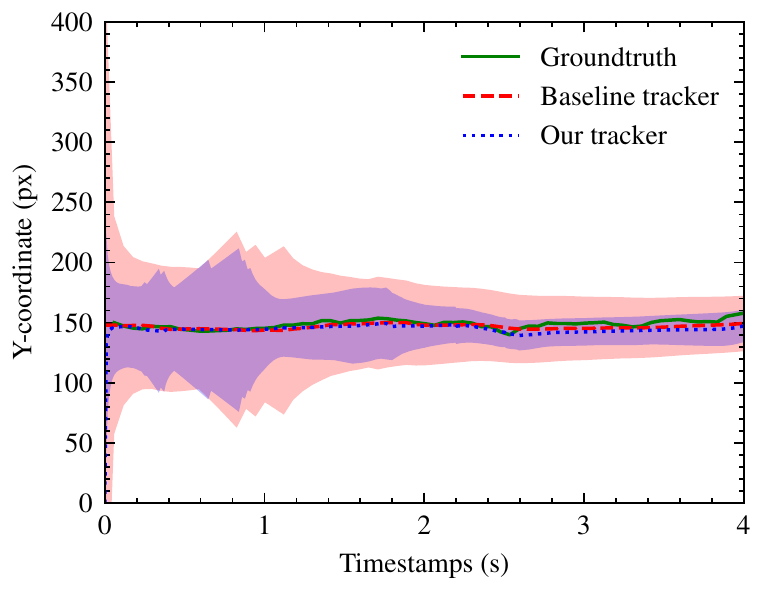}}%

    \caption{Estimate of the target position on a sequence with no motion blur. The covariance error is lower for our tracker, even though both produce a similar mean estimate.} 
    \label{fig:estim-person}
\end{figure}

First, we test both trackers on a sequence that shows no motion blur or other issues on the conventional frames. The results can be seen in Fig. \ref{fig:estim-person} (the estimates are plotted with three standard deviations) and Table \ref{tab:estim-error-cat}, which shows the estimation error values obtained with both options, computed according to \eqref{eq:error-cov} and \eqref{eq:error-x}. For this case, both trackers are able to similarly estimate the location of the target. However, due to the fact that reconstructed frames can be produced at higher frame-rates, the tracking uncertainty measured by the covariance of the estimation error is lower for our tracker, since obtaining more frequent measurements diminishes the uncertainty of the estimation.

\begin{figure}[t]
\centering
  \subfloat[X-Coordinate]{\includegraphics[width=0.43\textwidth]{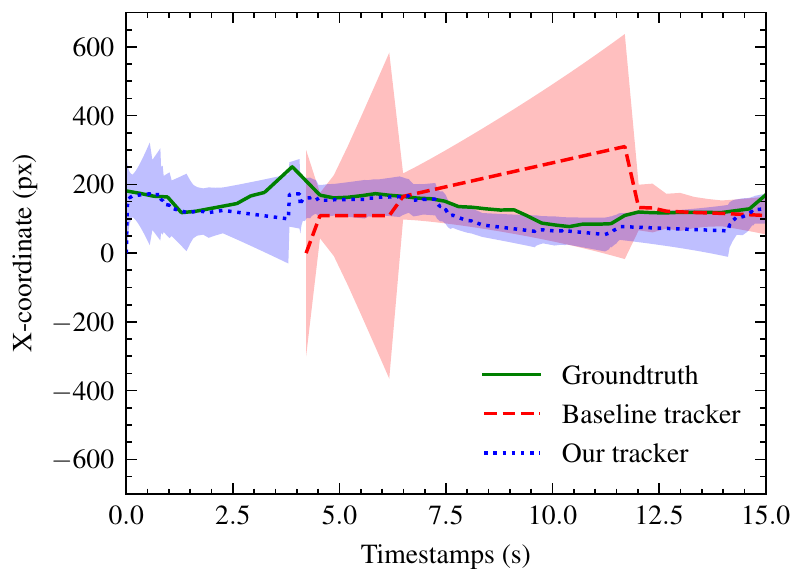}}\hfill%
  \subfloat[Y-Coordinate]{\includegraphics[width=0.43\textwidth]{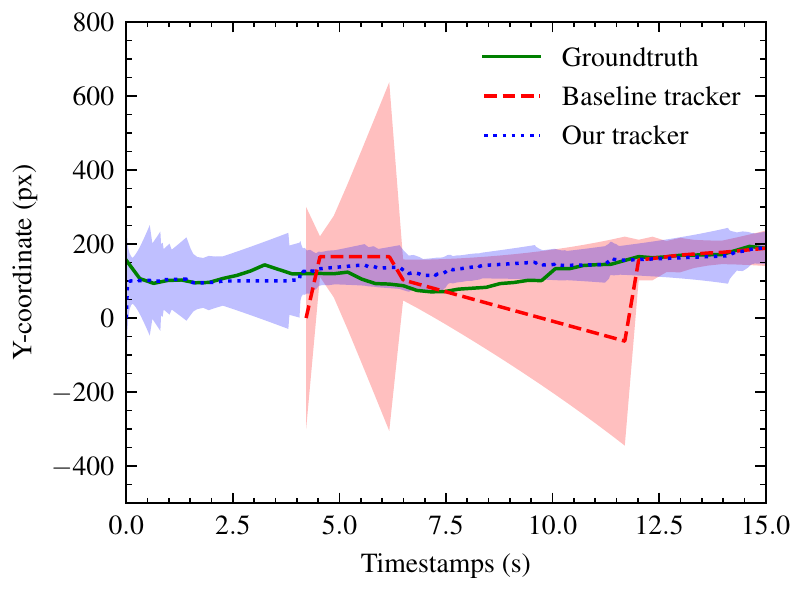}}%
     
    \caption{Estimate of the target position on a sequence with strong motion blur. Our tracker is able to produce a significantly more accurate estimate of the position of the target than the baseline tracker using conventional frames, showing that we successfully take advantage of the event camera for challenging visual conditions.} 
    \label{fig:estim-cat}
\end{figure}

\begin{table}[t]
\caption{Estimation errors for two sequences. When no motion blur is present, the error covariance is lower for our tracker, even though both trackers produce a similar mean estimate. For a sequence with strong motion blur, our tracker shows lower error values.}
\label{tab:estim-error-cat}
\centering
\begin{tabular}{ccccc}
\cline{2-5}
    & \multicolumn{2}{c}{No motion blur} & \multicolumn{2}{c}{Motion blur} \\ \cline{2-5} 
    & $\;\;$Baseline tracker$\;\;$   & $\;\;$Our tracker$\;\;$   & $\;\;$Baseline tracker$\;\;$  & $\;\;$Our tracker$\;\;$ \\ \hline
$\mathcal{E}_\text{x}$ (px)  & 56.05              & 48.17         & 114.02            & 45.81       \\ 
$\mathcal{E}_\text{gt}$ (px) & 14.64              & 17.35         & 145.84            & 51.33       \\ \hline
\end{tabular}
\end{table}

The usefulness of our event tracker is truly seen in sequences that are highly dynamic or have other attributes that make the images captured by a conventional camera unclear. For example, Fig. \ref{fig:estim-cat} shows the estimation results for a sequence with strong motion blur on the conventional frames, making detection impossible in most cases. This produces a poor estimate: since there are very few available measurements to correct the Kalman prediction, the filter relies too heavily on the process knowledge, and the covariance of the estimation error $\hat{P}$ increases greatly. Our tracker is able to produce a better estimate of the target location. Since the reconstructed frames provide a clearer image of the scene under these extreme motion conditions, we are able to obtain more measurements of the target's position, thus reducing the error and uncertainty of the estimation. 
Additionally, note that there is an initial period for the conventional frames where no measurements are being obtained due to the blurriness of the images, while the detector is able to locate the object in the reconstructed frames at that time.
Table \ref{tab:estim-error-cat} also includes the values of the evaluation metrics for this sequence, showing the better performance of our tracker versus the baseline. For both error metrics, our proposal achieves much smaller values of error. The metrics $\mathcal{E}_\text{x}(T_s,T_f)$ and $\mathcal{E}_\text{gt}(T_s,T_f)$ were computed from the time $T_s$ since the object is first detected in each case.

\subsection{Timing Performance}

So far, we have shown the advantages of using reconstructed frames from events in the context of visual object tracking. Another aspect to take into account is the additional latency that creating a reconstruction introduces. For the tracking experiments presented above it has been considered that, in order to estimate the state of the target at time $k$, the measurement $\mathbf{z}_k$ is available. Actually, obtaining the measurement comes with the temporal cost of creating a reconstruction and applying the detection network on it to locate the target. Since this time interval is constant for every measurement, the result is that all estimates are delayed a fixed amount of time. In our case, both the reconstruction and the detection task take about 10 ms to complete on a \textit{GeForce GTX 1080 Ti/PCIe/SSE2} GPU, resulting in a 20 ms delay from the moment that the event tensor is created.
The experiments have been done on pre-recorded sequences of events, so no limitation to the size of the event windows has been applied, other than using a value of $N$ that is appropriate for obtaining clear results. In an application with events being processed as the camera captures them, our tracker would be able to handle an incoming stream of events in intervals as small as 10 milliseconds, by running the reconstruction and detection networks in parallel on our GPU, producing output estimates of the location of the target at a frequency up to 100 Hz. However, these considerations are largely dependent on the hardware that is used. The performance could be improved by implementing the tracker on hardware optimized for inference using neural networks. 
Also, note that the event window size may need to be taken into account in a global comparison of latency between different approaches: methods that are applied directly to event data sometimes group events in larger windows (30 - 100 ms in \cite{Perot}, \cite{Iacono2018}), since the sparse event representation may not contain enough information in small windows.

\section{Conclusions}

We have presented a framework to breach the gap between events and deep learning for object detection and tracking applications. A detection-based object tracker that relies solely on event information has been implemented, by reconstructing images from events via an E2VID network and then performing detection with a YOLO neural network. It has been shown that this framework maintains the advantages of event cameras: the event tracker shows a superior performance to conventional frames in scenes where fast motions or challenging light conditions are present, where conventional cameras are not able to produce clear images for detection.

%

\bibliographystyle{styles/bibtex/splncs03_unsrt} 
\bibliography{bibliography} 

\begin{thebibliography}{10}
\providecommand{\url}[1]{\texttt{#1}}
\providecommand{\urlprefix}{URL }

\bibitem{gallego2022}
Gallego, G., Delbrück, T., Orchard, G., Bartolozzi, C., Taba, B., Censi, A.,
  Leutenegger, S., Davison, A.J., Conradt, J., Daniilidis, K., Scaramuzza, D.:
  Event-based vision: A survey. IEEE Transactions on Pattern Analysis and
  Machine Intelligence  44(1),  154--180 (2022)

\bibitem{Dubeau}
Dubeau, E., Garon, M., Debaque, B., Charette, R.d., Lalonde, J.F.: {RGB-D-E}:
  Event camera calibration for fast 6-{DOF} object tracking. In: 2020 IEEE
  International Symposium on Mixed and Augmented Reality (ISMAR). pp. 127--135
  (2020)

\bibitem{Zhang2021}
Zhang, J., Yang, X., Fu, Y., Wei, X., Yin, B., Dong, B.: Object tracking by
  jointly exploiting frame and event domain. In: IEEE/CVF International
  Conference on Computer Vision. pp. 13043--13052 (2021)

\bibitem{Wang2021}
Wang, X., Li, J., Zhu, L., Zhang, Z., Chen, Z., Li, X., Wang, Y., Tian, Y., Wu,
  F.: Vis{E}vent: Reliable object tracking via collaboration of frame and event
  flows. doi:arxiv-2108.05015  (2021)

\bibitem{Liu2016}
Liu, H., Moeys, D.P., Das, G., Neil, D., Liu, S.C., Delbrück, T.: Combined
  frame- and event-based detection and tracking. In: 2016 IEEE International
  Symposium on Circuits and Systems (ISCAS). pp. 2511--2514 (2016)

\bibitem{Jiang2021}
Jiang, R., Wang, Q., Shi, S., Mou, X., Chen, S.: {Flow-assisted visual tracking
  using event cameras}. CAAI Transactions on Intelligence Technology  6(2),
  192--202 (2021)

\bibitem{Perot}
Perot, E., de~Tournemire, P., Nitti, D., Masci, J., Sironi, A.: {Learning to
  Detect Objects with a 1 Megapixel Event Camera}. In: Advances in Neural
  Information Processing Systems. vol.~33, pp. 16639--16652 (2020)

\bibitem{Iacono2018}
Iacono, M., Weber, S., Glover, A., Bartolozzi, C.: {Towards Event-Driven Object
  Detection with Off-the-Shelf Deep Learning}. In: IEEE International
  Conference on Intelligent Robots and Systems. pp. 6277--6283. Institute of
  Electrical and Electronics Engineers Inc. (2018)

\bibitem{Jiang2020}
Jiang, R., Mou, X., Shi, S., Zhou, Y., Wang, Q., Dong, M., Chen, S.: {Object
  tracking on event cameras with offline–online learning}. CAAI Transactions
  on Intelligence Technology  5(3),  165--171 (2020)

\bibitem{Rebecq2019}
Rebecq, H., Ranftl, R., Koltun, V., Scaramuzza, D.: High speed and high dynamic
  range video with an event camera. IEEE Transactions on Pattern Analysis and
  Machine Intelligence  43(6),  1964--1980 (2021)

\bibitem{Scheerlinck2020}
Scheerlinck, C., Rebecq, H., Gehrig, D., Barnes, N., Mahony, R.E., Scaramuzza,
  D.: Fast image reconstruction with an event camera. In: 2020 IEEE Winter
  Conference on Applications of Computer Vision (WACV). pp. 156--163 (2020)

\bibitem{yolov5}
Jocher, G.: {YOLOv5n 'Nano' models, Roboflow integration, TensorFlow export,
  OpenCV DNN support}. doi:10.5281/zenodo.5563715  (2021)

\bibitem{Soderstrom2002}
S{\"{o}}derstr{\"{o}}m, T.: {Discrete-Time Stochastic Systems}. Springer-Verlag
  London (2002)

\bibitem{Kalman1960}
Kalman, R.E.: {A new approach to linear filtering and prediction problems}.
  Journal of Basic Engineering  82(1),  35--45 (1960)

\bibitem{Mueggler}
Mueggler, E., Rebecq, H., Gallego, G., Delbruck, T., Scaramuzza, D.: The
  {E}vent-{C}amera {D}ataset and {S}imulator: Event-based data for pose
  estimation, visual odometry, and {SLAM}. The International Journal of
  Robotics Research  36(2),  142--149 (2016)

\end{thebibliography}

\end{document}